\documentclass[letterpaper]{article} 
\usepackage{aaai25}  
\usepackage{times}  
\usepackage{helvet}  
\usepackage{courier}  
\usepackage[hyphens]{url}  
\usepackage{graphicx} 
\usepackage{natbib}  
\usepackage{caption} 
\usepackage{algorithm}

\usepackage{newfloat}
\usepackage{listings}
\DeclareCaptionStyle{ruled}{labelfont=normalfont,labelsep=colon,strut=off} 
\floatstyle{ruled}
\newfloat{listing}{tb}{lst}{}
\floatname{listing}{Listing}
\usepackage{amsmath,amsfonts,bm}

\def\eqref#1{equation~\ref{#1}}

\def\1{\bm{1}}

\DeclareMathAlphabet{\mathsfit}{\encodingdefault}{\sfdefault}{m}{sl}
\SetMathAlphabet{\mathsfit}{bold}{\encodingdefault}{\sfdefault}{bx}{n}

\def\gD{{\mathcal{D}}}

\def\gH{{\mathcal{H}}}
\def\gI{{\mathcal{I}}}

\def\gM{{\mathcal{M}}}

\def\gR{{\mathcal{R}}}

\def\gT{{\mathcal{T}}}

\def\gV{{\mathcal{V}}}

\DeclareMathAlphabet\mathbfcal{OMS}{cmsy}{b}{n}
\newcommand{\Sm}{\mathcal{S}}
\newcommand{\Am}{\mathcal{A}}
\newcommand{\Tm}{\mathcal{T}}
\newcommand{\Rm}{\mathcal{R}}

\newcommand{\alg}{CHiRP} 
 
\newcommand{\catrl}{CAT+RL}

\newcommand{\ssps}{$\mathbfcal{M}$}

\newcommand{\ssp}{$\mathcal{M}$}

\newcommand{\cat}{$\Delta$}

\newcommand{\prcat}{\Delta}

\newcommand{\Op}{\mathcal{O}}
\newcommand{\op}{o}

\newcommand{\si}{$s_{i}$}

\newcommand{\Sg}{$S_{g}$}

\newcommand{\absS}{\overline{\Sm}}
\newcommand{\abss}{\overline{s}}

\newcommand{\absSg}{\overline{S}_g}

\newcommand{\abstraj}{\overline{\tau}}
\newcommand{\traj}{\tau}

\newcommand{\picat}{$\pi$}

\newcommand{\pcats}{C-CATs}
\newcommand{\pcat}{C-CAT}
\usepackage{amsthm}
\usepackage[algo2e,linesnumbered,ruled,vlined]{algorithm2e}
\usepackage{amssymb}
\usepackage{algpseudocode}
\usepackage{amsmath,empheq}
\usepackage{xcolor}
\usepackage{subcaption}
\usepackage{scalerel}
\usepackage{booktabs}
\usepackage{eqparbox} 

\theoremstyle{definition}
\newtheorem{definition}{Definition}[section]

\title{Autonomous Option Invention for \\ Continual Hierarchical Reinforcement Learning and Planning}
\author{
    Rashmeet Kaur Nayyar {\normalfont and}
    Siddharth Srivastava
}
\affiliations{

    Autonomous Agents and Intelligent Robots Lab,\\
    School of Computing and Augmented Intelligence, Arizona State University, Tempe, AZ, USA\\

\{rmnayyar, siddharths\}@asu.edu
}

\usepackage{bibentry}

\begin{document}

\maketitle

%

\begin{abstract}
Abstraction is key to scaling up reinforcement learning (RL). However, autonomously learning abstract state and action representations to enable transfer and generalization remains a challenging open problem. This paper presents a novel approach for inventing, representing, and utilizing options, which represent temporally extended behaviors, in continual RL settings. Our approach addresses streams of stochastic problems characterized by
long horizons, sparse rewards, and unknown transition and reward functions. 

Our approach continually learns and maintains an interpretable state abstraction, and uses it to invent high-level options with abstract symbolic representations. These options meet three key desiderata: (1) composability for solving tasks effectively with lookahead planning, (2) reusability across problem instances for minimizing the need for relearning, and (3) mutual independence for reducing interference among options. 
Our main contributions are approaches for continually learning transferable, generalizable 
options with symbolic representations, and for integrating search techniques with RL to efficiently plan over these learned options to solve new problems. Empirical results demonstrate that the resulting approach effectively learns and transfers abstract knowledge across problem instances, achieving superior sample efficiency compared to state-of-the-art methods.
\end{abstract}

\section{Introduction} 
\label{introduction}

Reinforcement Learning (RL) for enabling autonomous decision-making has been constrained by two fundamental challenges: sample inefficiency and poor scalability, particularly in environments with long horizons and sparse rewards. To address these limitations, researchers have focused on reducing the problem complexity through: (1) state abstraction, which creates compact state representations \cite{jong2005state, dadvar2023conditional},
and (2) temporal abstraction,
which captures hierarchical task structures through temporally extended behaviors \citep{barto2003recent, pateria2021hierarchical}, such as options \citep{sutton1999between}.
Abstraction-based methodologies offer principled approaches for knowledge transfer across tasks \citep{abel2018state}, especially in the challenging setting of continual learning \citep{
liu2021lifelong, khetarpal2022towards},
where agents must interact with and solve tasks indefinitely. 
However, most existing research on option discovery in RL focuses either on continuous control tasks with short horizons and dense rewards \citep{bagaria2021skill, klissarov2021flexible}, on single-task settings \citep{bagaria2020option, riemer2018learning}, or lacks support for lookahead planning over options to guide low-level policy learning \citep{machado2017eigenoption, khetarpal2020options}. A critical challenge remains open: the autonomous discovery of generalizable, reusable options for long-horizon, sparse-reward tasks in continual RL settings.
This is particularly relevant to real-world scenarios such as warehouse management, disaster recovery operations, and assembly tasks, where agents must adapt to shifts in context and required behaviors without 
dense reward feedback and 
closed-form analytical models.

We present a novel approach that coherently addresses option discovery and transfer with a unified symbolic abstraction framework
for factored domains in the continual RL settings.
We focus on long-horizon, goal-based Markov decision processes
(MDPs) 
in RL settings with unknown transition functions and sparse rewards. In particular, we consider three conceptual desiderata for options: 1) \emph{composability}: support chaining options for enabling hierarchical planning, 2)
\emph{reusability}: support transfer of options to new problem instances, minimizing the need for relearning,
and 3) \emph{mutual independence}: reduce interference among options, 
allowing options to be learned and executed independently with minimal side effects while ensuring composability at well-defined endpoints. 
Most prior works meet some of these criteria but not all.

\begin{figure}[t!]
     \centering
    \includegraphics[scale=0.35]{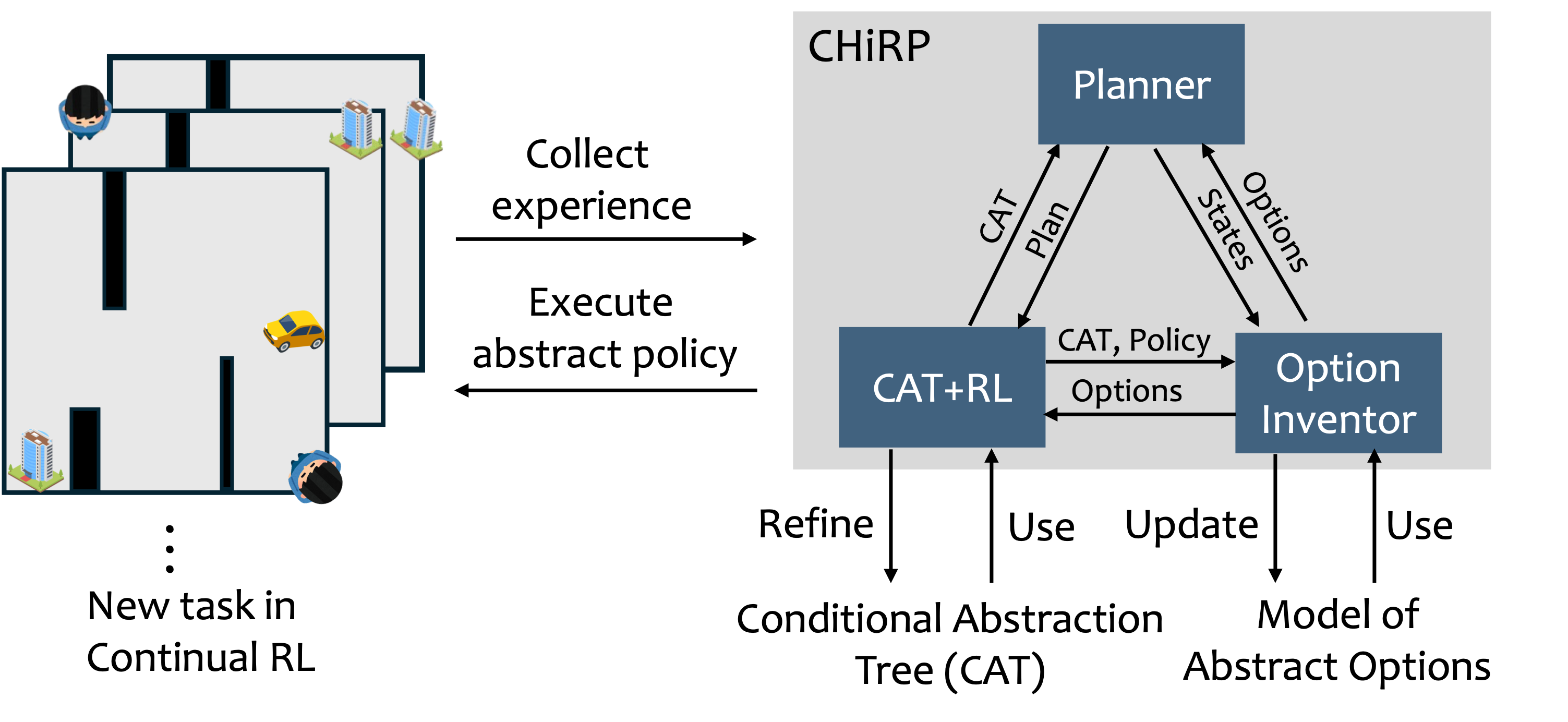}
\caption{\small 
Overall approach for Continual Hierarchical Reinforcement Learning and Planning (CHiRP).}

\label{fig:block}
\end{figure}

Our approach, \textbf{C}ontinual \textbf{Hi}erarchical \textbf{R}einforcement Learning and \textbf{P}lanning 
(\alg{})
takes as input a set of state variables and a stochastic simulator, and invents options satisfying desiderata 1-3:
the invented options have symbolic abstract descriptions that directly support composability and reusability through high-level planning;
these options have stronger effects on different sets of variables and/or values, supporting mutual independence.
The core idea is to capture notions of context-specific 
abstractions that depend on and change with the current state
by identifying salient variable values responsible for greatest variation in
the Q-function, and to use changes in 
these abstractions
as a cue for defining option endpoints.
With every new task in a continual stream of problems, \alg{} transfers these options and invents new options, building a model of options that is more broadly useful (Fig.~\ref{fig:block}). For example, in large instances of the well-known taxi domain \citep{dietterich2000hierarchical}, \alg{} 
autonomously invents four key options:
navigate to the passenger location, pickup the passenger, navigate to the dropoff location, and dropoff the passenger.

Extensive empirical evaluation across 
a variety of challenging domains with continuous/hybrid states and discrete actions
demonstrates that our approach substantially surpasses SOTA RL baselines in sample efficiency within continual RL settings.
Key strengths of our approach are: fewer hyper-parameters and less tuning required
compared to many of the baselines, including SOTA DRL methods that require extensive architecture tuning, 
greater interpretability, 
increased sample-efficiency, and satisfaction of key conceptual desiderata for task decomposition.

To the best of our knowledge, \alg{} is the first approach to 
autonomously invent composable, reusable, and mutually independent options 
using auto-generated state abstractions, and to use these options to create a novel hierarchical paradigm  
for continual RL in long-horizon, sparse reward settings.
Our main contributions are: 
(a) a novel approach for auto-inventing symbolic options with abstract representations,
(b) a novel search process for composing options for solving new tasks, and
(c) a hierarchical framework 
that integrates planning and learning
for continual RL.

\section{Formal Framework}

\paragraph{Problem Definition.}

We assume RL settings where an agent interacts with a goal-oriented Markov decision process (MDP)  \ssp{}
defined by the combination of an environment $\langle \Sm, \Am, \Tm, \gamma, h \rangle$ and a task $\langle$\si{}, \Sg{}, $\Rm \rangle$.
Here, $\Sm$ is a set of 
states defined using a factored representation, where $\mathcal{V} = \{v_1,\emph{\dots},v_n\}$ is a set of continuous real-valued or discrete variables, and each $v_i \in \mathcal{V}$ has an 
ordered domain $\gD_{v_i} = [\gD_{v_i}^\emph{min}, \gD_{v_i}^\emph{max})$. 
We denote the value of variable $v_i$ in state $s$ as $s(v_i)$. 
Each state $s \in \Sm$ is defined by assigning each $v_i \in \gV$ a value from its domain, i.e., $s(v_i) \in \gD_{v_i}$.
$\Am{}$ is a set of finite actions. $\Tm{}:\Sm{}\times\Am{} \rightarrow \mu \Sm$ is a stochastic transition function where $\mu \Sm$ is 
a probability measure on $\Sm$. 
$\gamma \in (0,1]$ is a discount factor and $h$ is a horizon. Lastly, \si{} $\in \Sm{}$ is an initial state, \Sg{} $\subseteq \Sm{}$ is a set of goal states, and $\Rm{}:\Sm{}\times\Am{} \rightarrow \mathbb{R}$ is a reward function underlying the task.

\paragraph{Running Example.} Consider a hybrid state space (defined by both continuous and finite variables) adaptation of the classic taxi domain \citep{dietterich1999state}, where a taxi starts at a random location and is tasked with picking up a passenger and transporting them to their destination. The pickup and dropoff locations are chosen randomly among $n$ specific locations. 
States are represented by variables $\gV{}$: 
$x \in \mathbb{R}$ (taxi's x-coordinate), 
$y \in \mathbb{R}$ (taxi's y-coordinate), 
$l \in \{0,1,\dots,n\}$ (passenger's location, where integers indicate specific locations and $0$ indicates elsewhere), and $p \in \{0,1\}$ (passenger's presence in the taxi). For clarity, consider variables with small domains: $\gD_{x} = [0.0, 5.0), \gD_{y} = [0.0, 5.0)$,
$\gD_{l} = \{0,1,2,3,4\}$, and $\gD_{p} = \{0,1\}$.
A state assigns a value to each variable from its domain, e.g., $s = \langle s(x) = 0.9, s(y) = 2.1, s(l) = 3, s(p) = 0 \rangle$.  
There are six primitive actions: navigate in four cardinal directions by a fixed distance, pickup the passenger, and drop-off the passenger.

A solution to a goal-oriented MDP is a policy $\pi:\Sm \rightarrow \Am$, which maps each state $s \in \Sm$ 
to an action $a \in \Am$, with the objective of maximizing the expected discounted cumulative reward. When analytical models of transition $\Tm$ and reward $\Rm$ functions are available, classical dynamic programming methods such as value iteration \cite{bellman1957dynamic} and policy iteration \cite{howard1960dynamic} are used to compute policies. However, in RL settings, $\gT(\Sm, \Am)$ and $\gR(\Sm, \Am)$ can be sampled but their closed-form analytical models are not available. Methods like 
Q-learning \citep{watkins1992q}, DQN \citep{mnih2013playing}, and PPO \citep{schulman2017proximal} are designed to learn policies directly from samples, but
they are often sample inefficient and struggle to scale when effective horizons are long \citep{laidlaw2023bridging} and rewards are sparse \citep{dadvar2023conditional}.


\paragraph{Continual Reinforcement Learning.}
Many challenging real-world scenarios are captured by continual or lifelong learning setting 
\citep{
ring1994continual, liu2021lifelong},
where an agent must interact with and solve a stream of related tasks, randomly sampled from a distribution,
over the course of its lifetime. In these tasks, subtle aspects of the initial state, goal states, transition function, and reward function 
change over time. The goal is to efficiently retain and reuse knowledge from previous experiences to solve new tasks.
We adapt the definition of continual learning \citep{khetarpal2022towards} to goal-oriented MDPs as follows.

\begin{definition}[Continual Reinforcement Learning (CRL)] 
CRL problem is a stream of $n$ MDPs \ssps{} where each MDP $\mathcal{M}\in\text{\ssps{}}$
shares $\langle \Sm{}, \Am{}, \gamma, h \rangle$ and may have distinct $\langle \Tm{}, s_{i}^\gM, S_{g}^{\gM}, \Rm^\mathcal{M} \rangle$.
An agent interacts with each $\gM_i \in $ \ssps{} for a maximum of $\gH$ timesteps in the
order $i = 1,\dots,n$.
\label{def:cpl}
\end{definition}

A solution to a CRL problem is a policy $\pi^\gM: \Sm \rightarrow \Am$ for each MDP $\gM$ in the problem stream \ssps{}.
The goal is to solve each $\gM \in$ \ssps{} while minimizing agent interactions and maximizing the expected discounted cumulative reward. In such challenging settings, abstraction techniques emerge as powerful tools for 
improving scalability and generalization in RL \cite{li2006towards}.

\begin{figure}[t!]
     \centering
    \includegraphics[scale=0.36]{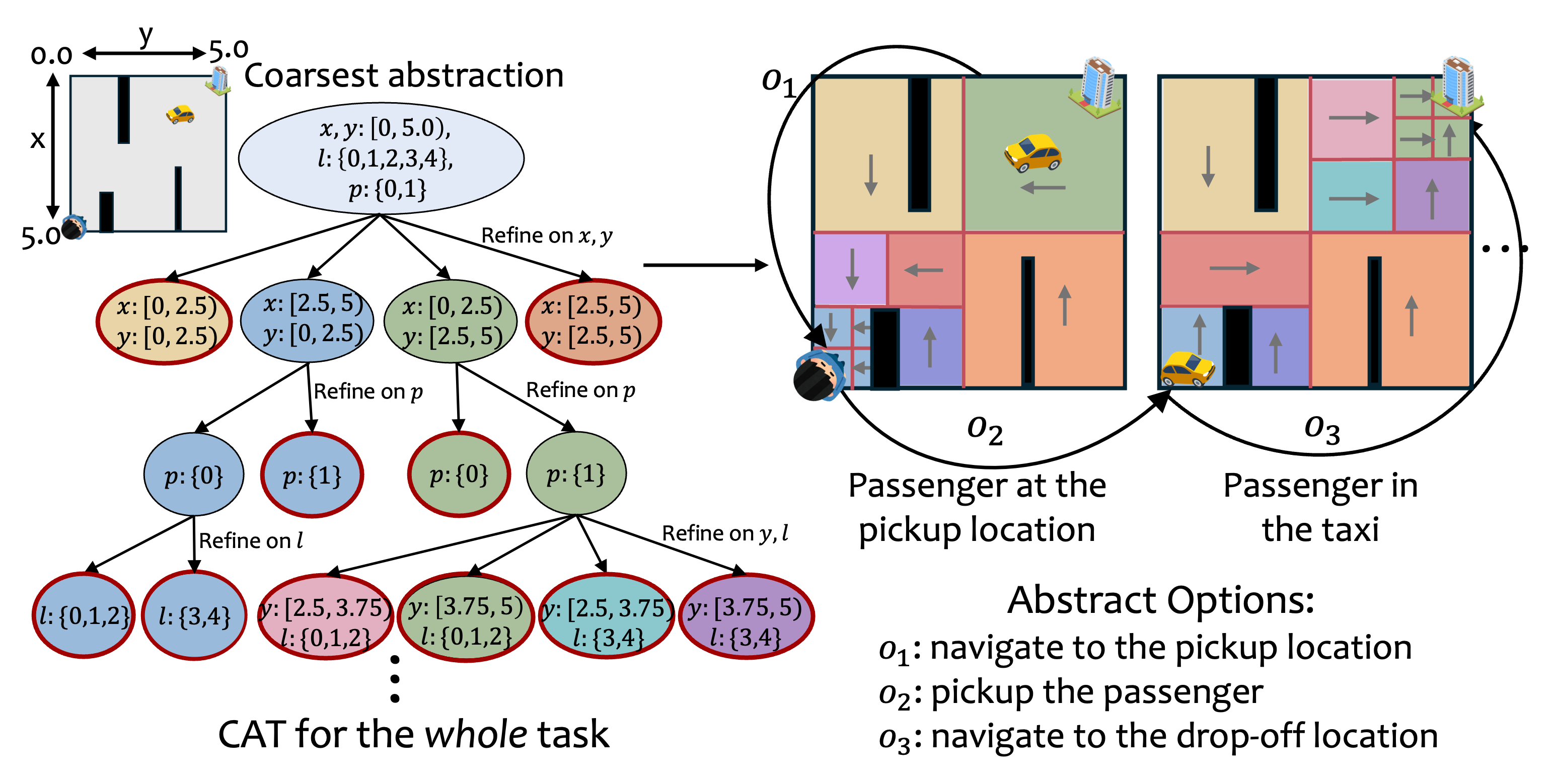}
\caption{\small 
Illustration of a Conditional Abstraction Tree (CAT) (left) and Abstract Options (right)
for a small instance in taxi world. Left: Nodes show values of refined variables; other variables inherit values from parent nodes.  
Right: Arrows denote option policies. Abstract states are highlighted with solid red lines in both figures.}

\label{fig:cat_to_options}
\end{figure}



\paragraph{State Abstractions.} 
A state abstraction $\phi:\Sm\rightarrow\absS$ maps each state $s \in \Sm$ to an abstract state $\abss \in \absS$, where $\absS$ is a partitioning of $\Sm$.  
Given a set of variables $\mathcal{V}$, let $\abss{}(v_i)$ denote the value of $v_i \in \mathcal{V}$ in an abstract state $\abss{}$.
An abstract state assigns an interval of values to each variable from its domain, e.g., 
state 
$s = \langle s(x) = 2.6, s(y) = 0.9, s(l) = 3, s(p) = 0 \rangle$
can be abstracted as 
$\abss = \langle \abss(x) = [2.5, 5), \abss(y) = [0.0, 2.5),  \abss(l) = \{3, 4\}, \abss(p) = \{0\} \rangle$. 
Here, $\abss{}$ represents a set of states
$\{s \in \Sm | \forall v_i \in \mathcal{V}$,  $s(v_i) \in \abss{}(v_i) \}$. 
Also, the coarsest state abstraction contains a single abstract state $\abss{}_{\emph{init}}$ that assigns $\abss{}_{\emph{init}}(v_i) = \gD_{v_i}$ to each $v_i \in \gV$. 
Formally, an abstract state is defined as follows.

\begin{definition}[Abstract State]
   Given a set of variables $\gV$ and the domain $\gD_{v_i}$ for each variable $v_i \in \gV$, an \emph{abstract state} $\abss{} \in \absS{}$ is defined by assigning an interval of values $\abss{}(v_i) \subseteq \gD_{v_i}$ 
   to each $v_i \in \gV$. 
\end{definition}

\paragraph{Conditional Abstraction Trees (CATs).} 
State abstraction on a variable’s values (such as taxi’s location) is \emph{conditioned} on the values of the other variables (such as passenger’s presence in the taxi). Such rich conditional abstractions for a task can be captured in the form of a Conditional Abstraction Tree (CAT) \citep{dadvar2023conditional} where the root denotes the coarsest abstract state, while lower-level nodes represent abstract states with greater refinement on variables requiring higher resolution in variable values. 
Fig.~\ref{fig:cat_to_options} illustrates a CAT for a small problem of taxi domain. CATs are defined formally as follows.

\begin{definition}[Conditional Abstraction Trees (CATs)]
A \emph{CAT} \cat{} is a tuple $\langle \mathcal{N}, \mathcal{E} \rangle$, where $\mathcal{N}$ is a set of nodes representing possible abstract states and $\mathcal{E}$ is a set of directed edges connecting these nodes. The root represents the coarsest abstract state $\abss{}_{\emph{init}}$. 
An edge $e \in \mathcal{E}$ from a parent abstract state $\abss{}_p \in \mathcal{N}$ to a child abstract state $\abss{}_c \in \mathcal{N}$ exists iff $\abss{}_c$ can be obtained by splitting atleast one of the variable intervals in $\abss{}_p$ at most once. 
The leaf nodes represent the 
active 
abstract state space $\absS{}_\text{\cat{}}$. 
\cat{} defines a state abstraction $\phi_\text{\cat{}}:\Sm{}\rightarrow \absS{}_\text{\cat{}}$ mapping each state $s \in \Sm{}$ to the abstract state $\abss{} \in \absS{}_\text{\cat{}}$ represented by the unique leaf in \cat{} containing $s$.
\end{definition}

In this work, we use novel notions based on CAT-based state abstractions to coherently address option invention and transfer with a unified abstraction framework. We compute CATs online using \catrl{} \citep{dadvar2023conditional} which refines states with high dispersion in TD-errors
during Q-learning over the abstract state space.

 \paragraph{Abstract Options.}
 We use the standard notion of options \citep{sutton1999between}. 
 An option $o$ is a triple $\langle \mathcal{I}_\op, \beta_\op, \pi_\op \rangle$, where $\mathcal{I}_\op \subset \Sm{}$ is the initiation set where $o$ can initiate, $\beta_\op \subset \Sm{}$ is the termination set where $o$ terminates, and $\pi_\op: \Sm{} \rightarrow \Am{}$ is the option policy prescribed by $o$ that maps states to actions. Our approach autonomously learns all components of options, defined over an abstract state space $\absS{}_{\text{\cat}_\op}$ 
as follows. We define an \emph{abstract option} $\op{}$ as a tuple $\langle \text{\cat{}}_\op, \mathcal{I}_\op, \beta_\op, \pi_\op \rangle$, where $\text{\cat{}}_\op$ is the CAT-based state abstraction  $\phi_\text{\cat{}$_\op$}: \Sm{} \rightarrow \absS{}_{\text{\cat}_\op}$, $\mathcal{I}_\op \subset \absS{}_{\text{\cat}_\op}$ is the 
abstract initiation set,
$\beta_\op \subset \absS{}_{\text{\cat}_\op}$ is the
abstract termination set,
and $\pi_\op:$ $\absS{}_{\text{\cat}_\op} \rightarrow \Am{}$ is the abstract partial policy. 
$\mathcal{I}_\op$ and $\beta_\op$ denote \emph{option endpoints}. The declarative description of an option is termed as \emph{option signature} $\langle \mathcal{I}_\op, \beta_\op \rangle$.
 Additionally, two options $\op_i$ and $\op_j$ are composable 
 iff  $\beta_{\op_i} \subseteq  \mathcal{I}_{\op_j}$.

\section{Continual Hierarchical RL and Planning}
\label{sec:our_approach}

The core contribution of this paper is a novel approach for autonomously inventing a forward model of abstract options using auto-generated CAT-based state abstractions and efficiently utilizing them for solving continual RL problems. Our approach \alg{}
(Fig.~\ref{fig:block})
takes as input a continual stream of tasks \ssps{} and a stochastic simulator, and computes a policy for each task. 
The key insight for option invention is that CATs, auto-generated using \catrl{}, for each task inherently capture abstractions that remain stable within a subtask, but change significantly across subtasks within the task.
We use CATs to
capture notions of context-specific 
abstractions that depend on the current state
and then use changes in these 
salient
abstractions
as a cue for defining option endpoints. For instance, in the taxi domain (Fig.~\ref{fig:cat_to_options}), when the passenger has not been picked up, the abstraction needs greater refinement on the value of the taxi's location closer to the passenger's location. However, when the context changes to a situation where the passenger is in the taxi and has not been dropped off, the abstraction needs greater refinement on the value of the taxi's location near the destination. In this scenario, the pickup option (option $o_2$ in Fig.~\ref{fig:cat_to_options}) can be seen as an option that achieves a significant change in 
context-specific abstractions.

\alg{} maintains a universal CAT created from the current and all previous problems in the stream, and exploits its structure to identify context-variables, such as the passenger's presence in the taxi. These variables are used to define a context-specific distance between states in a manner such that higher distances correspond to greater changes in salient variables and values. 
\alg{}
operationalizes this notion of changes in saliency to invent abstract options.
Note that the descriptions of the invented options are symbolic, hence directly support efficient composition and reuse. When a new task is encountered, \alg{} uses foresight to plan ahead by connecting endpoints of learned options, while also inventing additional option signatures to bridge gaps if needed. Each option maintains its own encapsulated CAT-based state abstraction, allowing options to be used and updated independently. The options have stronger effects on different sets of variables and values, which reduces mutual interference. 

In Sec.~\ref{sec:overview}, we present \alg{}, our overall approach to continual RL through autonomous invention, transfer, and reuse of abstract options. Sec.~\ref{sec:invent_options} details our novel approach for option invention, and
Sec.~\ref{sec:continual_rl} explains a novel planner for composing these options to solve new tasks.

\subsection{Algorithm Overview}
\label{sec:overview}

Given a continual RL problem, Alg.~\ref{alg:main} begins with an empty model of abstract options $\Op$ and a CAT \cat{} with the coarsest state abstraction (lines 1-2). For each new task in the stream, 
the CAT's abstraction is used to compute the initial and goal abstract states (line 4). Once a solution policy is found or a budget of $\gH$ timesteps is reached, the agent moves to solving the next task (line 5). To solve the current task, the agent interleaves: (1) a planner to plan with the current model of abstract options, (2) \catrl{} to refine the current CAT's state abstraction during learning, and (3) an option inventor to invent novel abstract options using the updated CAT (lines 6-16). 
The updated model of options $\Op$ and CAT \cat{} are transferred to solve subsequent tasks (line 3).

Given a new task, Alg.~\ref{alg:main} uses an offline search process with the current model of abstract options to compute a plan from the current abstract state to a goal abstract state, denoted by $\Pi{} = \langle o_0,\dots,o_n \rangle$, $o_i \in \Op{}$. The learned option representations are used to compose 
 this plan, as detailed in Sec.~\ref{sec:continual_rl} (line 6 computeOptionPlan()). The method additionally creates new option signatures to allow connecting endpoints of learned options with gaps between them. If no plan is found with the current model, we initialize the plan with a new option signature from the current abstract state to the goal abstract states (line 8). The CAT and policies for these options are learned later during RL.

 For each newly created option signature or previously learned option in the plan, $\op \in \Pi$, we generate an MDP with a sparse intrinsic reward for reaching the option's termination (line 10). The option's CAT $\text{\cat}_\op$ and policy $\pi_\op$ are then learned or fine-tuned using \catrl{} (line 11). We use these option-specific CATs and policies
to invent new 
abstract options with updated representations, as detailed in Sec.~\ref{sec:invent_options} (line 13 inventOptions()).  
This process converts option signatures into abstract options with learned CATs and policies.
We also update the universal CAT \cat{} with each invented option's CAT $\text{\cat}_\op$, adjusting the current abstract state (line 14). Note that option executions can be stochastic, and the agent may fail to successfully reach the termination set of an option. In such cases, we use active replanning \cite{kaelbling2011hierarchical} from the current abstract state to a goal abstract state (line 16), and continue learning option-specific CATs and policies. This process repeats until the computed plan of abstract options successfully solves the problem. Finally, Alg.~\ref{alg:main} transfers the updated model $\Op$ with the new options and CAT \cat{} to solve new tasks.

\SetAlCapFnt{\small}

\begin{algorithm}[t]
\caption{CHiRP algorithm}
\label{alg:main}
\SetKwComment{Comment}{/* }{ */}

\KwIn{Stream of MDPs \ssps{}, Budget $\gH$
}
\KwOut{Policy $\pi^\gM$ for each $\gM \in\text{\ssps}$ }

$\Op{} \leftarrow$ Initialize empty model of abstract options

\cat{} $\leftarrow$ Initialize CAT with $\abss{}_{\emph{init}}$

\For{$\gM$ $\in$ \ssps{}}{

    $s$ $\leftarrow$ $s_{i}^{\gM}$;  $\abss{}$, $\absSg{}$ $\leftarrow$ abstractStates(\cat{}, $s_{i}^{\gM}$, $S_{g}^{\gM})$ 
    
        \While{$\pi^\gM$ not found \text{or} steps $< \gH$}{

      $\Pi$ $\leftarrow$  computeOptionPlan($\Op{}$, \cat{}, $\abss{}$, $\absSg{}$)
      
     \If{$\Pi$ is not found}{
            $\Pi$ $\leftarrow$ inventOptionSign($s$, $S_{g}^{\gM}$) 
        }

        \For{$\op{} \in$ $\Pi$ }{ 
              \ssp{}$_o$ $\leftarrow$ generate MDP for $\op{}$
              
             \cat{}$_\op$, \picat{}$_\op$  $\leftarrow$ \catrl{}(\ssp{}$_o$, \cat{}, $s$)
             
            \If{\picat{}$_\op$ is learned}{
                 $\Op.$update$($inventOptions$($\cat{}$_\op$$, $\picat{}$_\op))$
                 
                 \cat{} $\leftarrow$ \cat{}$_\op$; update $s, \abss$  
            }
            \Else{
                break and replan $\Pi$ 
            }

        }
    }

}
      \KwRet $\forall \gM \in\text{\ssps}$ $ \pi^\gM$

\end{algorithm}

\subsection{Inventing Generalizable Options}
\label{sec:invent_options}

\begin{algorithm}[t]
\caption{Invention of Abstract Options}
\label{alg:invent_options}

\SetKwComment{Comment}{/* }{ */}
\setcounter{AlgoLine}{0}

\KwIn{
CAT \cat{}, Policy \picat{}, thresholds $\delta_{\emph{thre}}$ and $\sigma_{\emph{thre}}$
}

\KwOut{Abstract options $\Op{}$}

 $\traj{} \leftarrow$ computeTrajectory(\picat{}) 

 $\abstraj{} \leftarrow$ computeAbstractTrajectory($\traj$,\cat) 


\cat$_\traj{} \leftarrow$ generateContext-SpecificCATs(\cat{},$\traj{}$) 

$\traj{}^* \leftarrow$ identifyOptionEndPoints(\cat{},\cat$_\traj{}$,$\abstraj$,$\delta_{\emph{thre}}$,$\sigma_{\emph{thre}}$) 

$\Op \leftarrow$ inventAbstractOptions($\traj{}^*$,$\abstraj$,\cat{},$\pi$)

$\Op \leftarrow$ finetunePolicies($\Op$) 

\KwRet $\Op{}$ 
\end{algorithm}

We now discuss our approach for inventing options using a learned CAT and an abstract policy (Alg.~\ref{alg:main} line 13) (Sec.~\ref{sec:overview} explains our approach for obtaining these inputs). The key idea is to identify transitions that lead to significant changes in context-specific abstractions, revealing that the nature of the task has changed. We recognize changes in sets of salient variables and values that significantly impact changes in the CAT's structure, and use this as an indicator for determining when to initiate and terminate options. We also exploit the structure of different subtrees in the CAT to further decompose these options into multiple options. 

We first describe our overall approach for option invention (Alg.~\ref{alg:invent_options}) and then discuss each component in detail. Alg.~\ref{alg:invent_options} uses the input abstract policy \picat{} to compute a roll-out trajectory $\tau = \langle s_0, \dots, s_n \rangle$ and then uses the input CAT's abstraction function to compute its corresponding abstract trajectory $\abstraj$ (lines 1-2). Context-specific abstractions are generated for each state in the trajectory (line 3) and used to identify a sequence of option endpoints $\tau^*$ (line 4).
We use these identified option endpoints to invent abstract options as follows.
For each consecutive option endpoints $s^*_i, s^*_j \in \tau^*$, we first compute a segment $\text{seg}_{ij}= \langle \bar{s}_k,\dots,\bar{s}_m \rangle \subseteq \abstraj$  s.t. $\bar{s}_k = s^*_i$ and $\bar{s}_{\emph{m+1}} = s^*_j$. 
Then, an abstract option $o_{ij} = \langle \text{\cat{}}_{\op_{ij}}, \mathcal{I}_{\op_{ij}}, \beta_{\op_{ij}}, \pi_{\op_{ij}} \rangle$ is invented, where 
$\gI_{o_{ij}}$ contains siblings of $s^*_i$ in the CAT \cat{} that are in $\text{seg}_{ij}$, $\beta_{o_{ij}} = \{s^*_j\}$, \cat{}$_{o_{ij}}=$ \cat{}, and $\pi_{o_{ij}} = \pi$ (line 5). Finally, we fine-tune the policy for each invented option using \catrl{} (line 6). 
Options invented in this fashion are used to update the model of abstract options $\Op{}$. We now discuss our method for generating context-specific abstractions  (line 3) and identifying option endpoints (ine 4) in detail below.

\paragraph{Generating Context-Specific Abstractions.}
Recall that the learned CAT captures abstractions for the whole task.
We capture context-specific abstractions that depend upon the current state in the form of Context-Specific CATs (C-CATs), which are ablations of the input CAT. To generate C-CATs, we first identify \emph{context-variables} as variables that are relatively more refined among the ones that have a low frequency of change (\citet{hengst2002discovering}).
The degree of refinement for a variable  intuitively captures how close the abstraction is to its concrete representation. It is inversely proportional to the measure of reachable concrete values within each value interval of that variable. A high degree of refinement in a variable indicates that it significantly contributes to variation in the Q-function during learning of the CAT. 
For instance, in the taxi domain, variable $p$ denoting passenger's presence in the taxi is a context-variable since its value persists more and is more refined. 
The refinement of other frequently changing variables, 
such as taxi's location, depend on the values of the context-variables.  
We condition the input CAT on these context-variables to generate C-CATs.

\setlength{\abovecaptionskip}{0pt}
\setlength{\belowcaptionskip}{-12pt}
\begin{figure}
     \centering
\includegraphics[scale=0.32]{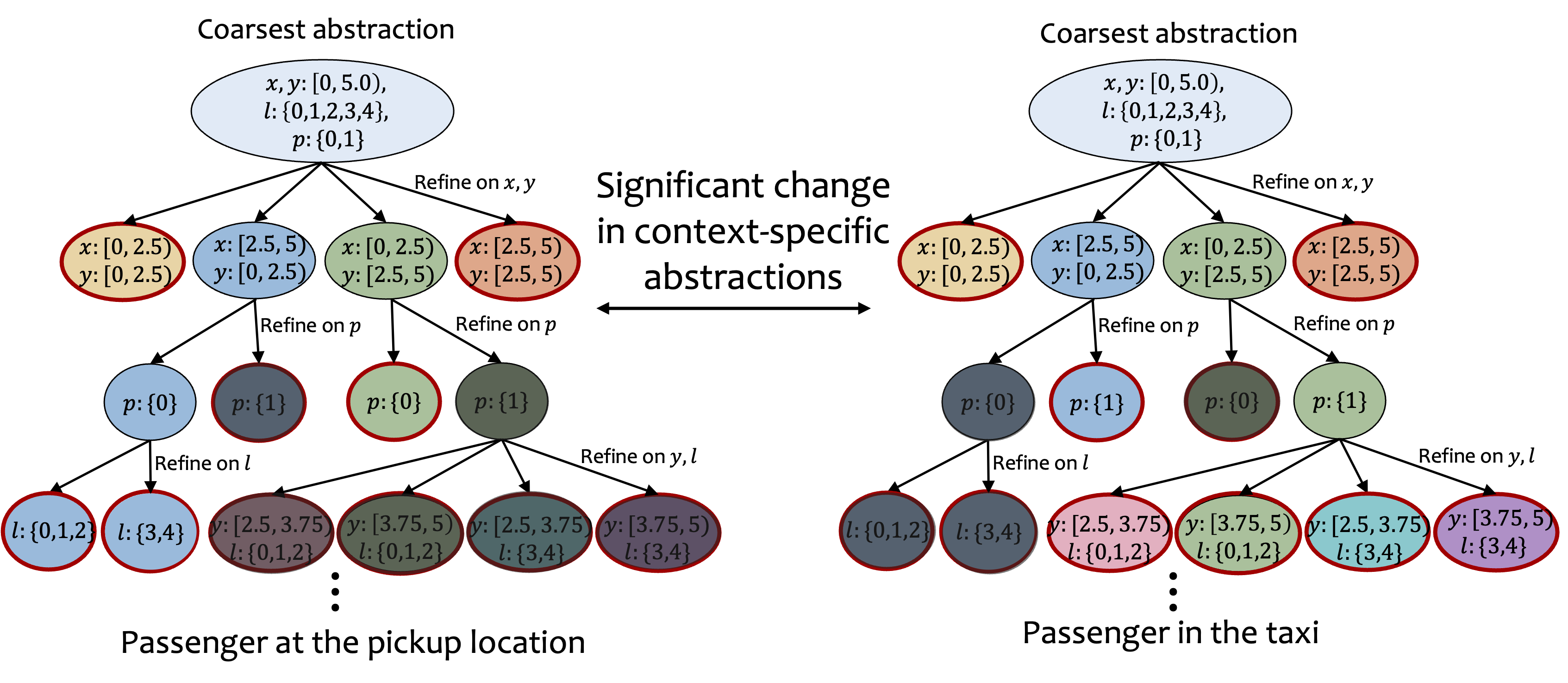}
\caption{\small
Illustration of two Context-Specific CATs (\pcats{}) highlighting different active abstractions (represented by leaves) in the CAT from Fig.~\ref{fig:cat_to_options}. The left \pcat{} corresponds to $p=0$, while the right \pcat{} corresponds to $p=1$. 
}
\label{fig:pruned_cats}
\end{figure}

Conditioning the CAT on context-variables highlights abstractions that are salient for different contexts\textemdash expressed by different ``active'' subtrees\textemdash within the CAT.  
More specifically, \pcats{} fix values of context-variables in the current state and preserve all abstract states in the CAT that are consistent with these fixed values. All other abstract states that are inconsistent are ignored (shown in dark in Fig.~\ref{fig:pruned_cats}).
For example, 
Fig.~\ref{fig:pruned_cats} (left) illustrates the \pcat{} for state $s_1$ 
where the passenger’s location is fixed at the bottom left and the passenger has not yet been picked up, i.e., $s_1(p)=0$. Similarly, Fig.~\ref{fig:pruned_cats} (right)  
illustrates the \pcat{} for state $s_2$ where the passenger has been picked up, i.e., $s_2(p)=1$. 
We formally define \pcats{} as follows.
 \begin{definition}[Context-specific CATs (\pcats{})]
Given a CAT \cat{} $= \langle \mathcal{N}, \mathcal{E} \rangle$ and context-variables $V \subseteq \mathcal{V}$, a \emph{\pcat{}} $\prcat{}_{s}$ for state $s$
is defined as $\langle \mathcal{N}', \mathcal{E}' \rangle$ where $\mathcal{N}' \subseteq \mathcal{N}$  s.t. $\mathcal{N'} = \{\abss{} | \abss{} \in \mathcal{N}, v_i \in V, s(v_i) \in \abss{}(v_i) \}$
and $\mathcal{E}' \subseteq \mathcal{E}$ s.t. $\mathcal{E}' = \{(\abss_1, \abss_2)| (\abss_1, \abss_2) \in \mathcal{E}, \abss_1, \abss_2 \in \mathcal{N}'\}$.  
\label{def:\pcats{}}
\end{definition}

\paragraph{Identifying Option Endpoints.} We identify transitions that lead to significant changes in context-specific abstractions to define endpoints of new options. 
For instance, consider \pcats{} in Fig.~\ref{fig:cat_to_options} before and after the passenger is picked up. These \pcats{}
 are significantly different from each other, indicating a significant change in abstraction. 
To measure
 difference between abstraction functions represented by two \pcats{} generated from the same CAT, we use a context-specific distance function. Intuitively, this distance is computed by traversing from the root node and summing the structural differences between corresponding subtrees of \pcats{}.
Given a \pcat{} $\prcat{}_s$ for state s, let 
 $\prcat{}_s^{n}$ denote the subtree rooted at node $n$, and 
  $\text{depth}_{\emph{max}}({\prcat{}_s^{n}})$ 
 denote the maximum depth of that subtree.
 We drop $n$ from $\prcat{}_s^{n}$ when $n=s_{init}$.
Let $n_i$ denote the $i$th child of node $n$ in the CAT \cat. 
We formally define this distance as follows.

\begin{definition}[Context-specific distance between \pcats{}]
    Given two \pcats{} obtained from \cat{} rooted at node $n$, $\prcat{}^{n}_{s_1}$ and $\prcat{}^{n}_{s_2}$, the distance between them is defined as
    \begin{empheq}[left={\delta(\prcat{}^{n}_{s_1},\prcat{}^{n}_{s_2})=\empheqlbrace}]{align*}
 & \text{depth}_{\emph{max}}(\prcat{}^{n}_{s_1}),\quad\text{if $n \text{ not in } \prcat{}^{n}_{s_2}$}; \\
 & \text{depth}_{\emph{max}}(\prcat{}^{n}_{s_2}),\quad\text{if $n \text{ not in } \prcat{}^{n}_{s_1}$}; \\
 & \Sigma_i \delta(\prcat{}^{n_i}_{s_1},\prcat{}^{n_i}_{s_2}),\quad\text{otherwise}.
\end{empheq}
\label{def:abstraction_dist}
\end{definition}

To identify option endpoints using trajectory $\tau$ and the context-specific distance, 
we first identify context-variables from CAT \cat{}, and generate \pcats{} \cat$_\traj{} = \langle \prcat{}_{s_0},\dots, \prcat{}_{s_n} \rangle$.
Let $\delta_{\emph{thre}}$ be a distance threshold.
Then, for each transition $(s_\emph{i}, s_{\emph{i+1}}) \subseteq \tau$, we use abstract states
$\phi_\text{\cat{}}(s_\emph{i})$ and $ \phi_\text{\cat{}}(s_{\emph{i+1}})$
to define option endpoints if $\delta(\prcat{}_{s_i}, \prcat{}_{s_{\emph{i+1}}}) > \delta_{\emph{thre}}$. The initial and goal abstract states are also included.

Additionally, our approach uses a context-independent distance, also derived from the CAT, to allow decomposing an option into multiple options. 
For example, navigating to the pickup location can be decomposed into first reaching the larger bottom-left quadrant and then the exact pickup location (Fig.~\ref{fig:cat_to_options}). Intuitively, this distance is greater between states 
that belong to highly distinct (having higher lowest common ancestor (LCA)) and highly refined CAT subtrees. 
Let $\text{depth}_{\emph{max}}$ be the maximum depth of the CAT, and  $\text{depth}(n_1,n_2)$ denote the number of edges between nodes $n_1$ and $n_2$. We formally define this distance as follows.

\begin{definition}[Context-independent distance between abstract states]
Given a CAT \cat{} and LCA of two abstract states $\abss_{1}$ and $\abss_{2}$, 
    the distance between them $\sigma_\text{\cat{}}(\abss_{1}, \abss_{2})$ is defined as the weighted sum of 
    ($\text{depth}_{\emph{max}} - \text{depth}(\text{root}, \emph{LCA}) + 1)$ and $(\text{depth}(\emph{LCA}, \abss_{1})$ + $\text{depth}(\emph{LCA}, \abss_{2}))/2$.
\label{def:abstract_state_dist}
\end{definition}

We decompose options by extending the previously computed sequence of option endpoints as follows. We compute trajectory segments $\traj{}_\emph{seg} = \langle s_k,\dots,s_m\rangle \subseteq \traj{}$ s.t. $\phi_\text{\cat}(s_k)$ and $\phi_\text{\cat}(s_m)$ are consecutive option endpoints. We also compute corresponding abstract trajectory segments $\abstraj{}_\emph{seg}$ using the CAT. Let   $\sigma_\emph{thre} \leq 1.0$ be a distance threshold and $\sigma_{\emph{max}}$ be the maximum  distance between abstract states in any transition in $\abstraj{}_\emph{seg}$. Then, for each transition $(\abss_{\emph{i-1}}, \abss_{\emph{i}}) \subseteq \abstraj{}_\emph{seg}$, $\abss_{i}$ is additionally identified as an option endpoint if $\sigma_\Delta(s_{\emph{i-1}},s_{i}) > \sigma_{\emph{thre}} \times \sigma_{\emph{max}}$.

\subsection{Planning over Auto-Invented Options}
\label{sec:continual_rl}

We now describe a novel planning process to compute a plan $\Pi{}$ for a new task using the learned model of abstract options and the learned CAT overlayed with abstract transitions, termed as a Plannable-CAT (Alg.~\ref{alg:main} line 6). 
This plannable-CAT is used to guide the search process over the option endpoints, while creating new option signatures to connect them when needed. We apply single-outcome determinization \citep{yoon2007ff} for planning by considering only the most likely effects (here, the termination sets) of options. 
To compute a plan of options, we first augment the Plannable-CAT with transitions between option endpoints, including lifted transitions between higher levels of abstract states. We then use A$^*$ search over this Plannable-CAT with a cost function that prioritizes lower-level transitions and a heuristic defined by the context-independent distance (Def.~\ref{def:abstract_state_dist}). The idea is to compose abstract transitions at different levels of abstractions. The resulting plan is refined by replacing consecutive higher-level transitions with a new option signature. This helps to bridge gaps between option endpoints. Finally, \alg{} interleaves the execution of the computed plan with learning of option policies for any newly created option signatures to solve the new task.

\captionsetup{font=small,skip=0pt}
\begin{figure*}[t!]
     \centering
     \small
         \includegraphics[scale=0.8]{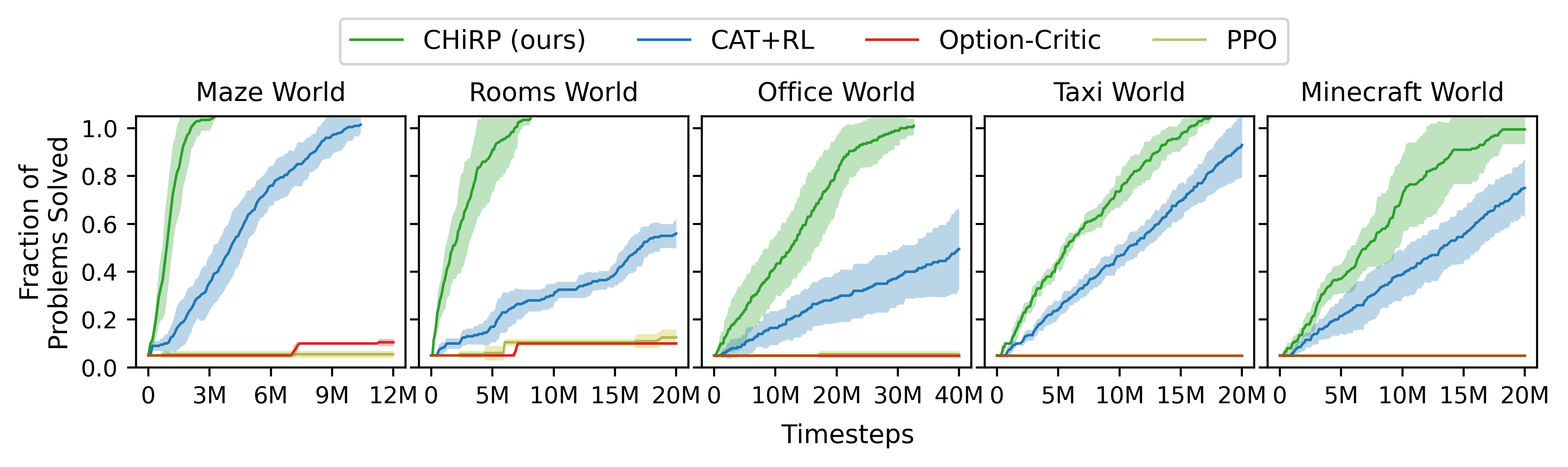}
        \caption{\small
        Fraction of tasks solved vs training steps, averaged over 10 independent trials. Each approach was evaluated on a sequence of 20 randomly sampled tasks in a continual learning setting, with a fixed budget of timesteps to solve each task. The timesteps include all environment interactions used for learning both abstractions and policies.
        }
        \label{fig:result}
\end{figure*}

\section{Empirical Evaluation}
\label{sec:empirical}

We evaluated \alg{}\footnote[1]{\small{https://github.com/AAIR-lab/CHiRP}} on a
diverse suite of challenging domains 
in continual RL setting. 
Full details about the used domains and hyperparameters are provided in the extended version of our paper \citep{rkn_aaai25}.

\subsection{Experimental Setup}
\paragraph{Domains.} 
For our evaluation, we compiled a suite of test domains for continual RL that are amenable to hierarchichal decomposition and challenging for SOTA methods. We then created versions of these problems that are significantly larger than prior investigations to evaluate whether the presented approaches are able to  push the limits of  scope and scalability of continual RL. 
Our investigation focused on \textit{stochastic} versions of the following domains with continuous or hybrid states: 
    (1)
    \textbf{Maze World} \citep{ramesh2019successor}: An agent needs to navigate through randomly placed wall obstacles to reach the goal; 
    (2)
    \textbf{Four Rooms World} \citep{sutton1999between}: An agent must move within and between rooms via hallways to reach the goal;
    (3)
    \textbf{Office World} \citep{icarte2018using}: An agent needs to collect coffee and mail from different rooms and deliver them to an office;
    (4)
    \textbf{Taxi World} \cite{dietterich2000hierarchical}: A taxi needs to pick up a passenger from its pickup location and drop them off at their destination;
     (5)
     \textbf{Minecraft} \cite{james2022autonomous}:
     An agent 
     must find and mine relevant resources, build intermediate tools, and use them to craft an iron or stone axe.

\paragraph{Baselines.} We selected the best-performing contemporary methods that do not require any hand-engineered abstractions or action hierarchies as baselines to match the absence of such requirements in our approach:
    (1)
    \textbf{Option-Critic} \citep{bacon2017option} is an end-to-end gradient-based method that learns and transfers option policies and termination conditions;
    (2)
    \textbf{\catrl{}} \citep{dadvar2023conditional} is a top-to-down abstraction refinement method that dynamically learns state abstractions during RL; and 
    (3)
    \textbf{PPO} \citep{schulman2017proximal} is a policy-gradient Deep RL method that progressively learns latent state abstractions through neural network layers. 

\paragraph{Hyperparameters.} 
A key strength of \alg{} over baselines is that it requires only five additional hyperparameters beyond standard RL parameters (e.g., decay, learning rate), unlike SOTA DRL methods that need extensive tuning and significant effort in network architecture design.
Throughout our experiments, we intuitively set $\delta_{\emph{thre}} = 0$ and $\sigma_{\emph{thre}} \sim 1$ to minimize hyperparameter tuning. These values are robust across domains,
preventing options from being too small or numerous. We use a limited set of values for $k_\emph{cap}$, $s_\emph{factor}$, and $e_\emph{max}$ parameters across different domains to adaptively control the training of an option's policy and CAT. All parameters are set to the same values across a continual stream of tasks.
Details on the used 
hyperparameters for \alg{} and the baselines are provided in the extended version.

\paragraph{Evaluation setting and metrics.} We evaluate in a continual RL setting where an agent needs to adapt to changes in initial states, goal states, transition and reward functions. For each domain, 20 tasks are randomly sampled sequentially from a distribution. Each approach is provided a fixed budget of $\gH$ timesteps per task before moving on to the next task. 
Due to stochasticity and lack of transition models, 
a task is considered solved if the agent achieves the goal  $\geq 90\%$ of the time among 100 independent evaluation runs of the learned policy. 
We report the fraction of tasks solved within the total allocated timesteps for each approach. The reported timesteps include all the interactions with the environment used for learning state abstractions, option endpoints, and option policies.  
Results are averaged, and standard deviations are computed from 10 independent trials across the entire problem stream. 

\subsection{Results}
We evaluate the presented work across a few key dimensions: sample-efficiency in continual RL setting, and satisfaction of key conceptual desiderata for task decomposition\textemdash composability, reusability, and mutual independence. 

\vspace{0.3em}
\textbf{Q1.}
\textit{Does \alg{} help improve sample-efficiency over SOTA RL methods in continual RL setting?} 

Fig.~\ref{fig:result} shows that \alg{} consistently outperforms all baselines. Our results confirm that, while in principle baseline approaches can solve problems without hand-designed abstractions and hierarchies, they require orders of magnitude more data and struggle to solve 
streams of distinct long-horizon tasks with sparse rewards.
We found that \catrl{} delivered the second-best performance, while Option-Critic and PPO consistently underperformed across all domains, failing to solve tasks within the allotted budget. While Option-Critic has the advantage of reusing options, it struggled to learn useful and diverse options.
 This is at least partly due to lack of mechanisms for modelling initiation sets for options and inability to plan long-term sequences of options,
leading to initiation of options in states where they were either ineffective or unnecessary. \catrl{} performed well by learning appropriate state abstractions for each task but did not 
decompose these tasks into transferable options. 
PPO struggled to learn likely due to the challenges associated with learning in environments with longer effective horizons and sparse rewards, as shown by \citet{laidlaw2023bridging}. Gradient-based methods typically rely on dense reward shaping for local gradient updates. In contrast, \alg{} overcomes these limitations by learning options with limited, symbolically represented initiation sets, 
 maintaining option policies at different levels of
 abstraction, and performing long-term planning with these options. \alg{} benefits from auto-generated state abstractions and goes beyond by inventing reusable options, resulting in more effective generalization and transfer across tasks.

\vspace{0.3em}
\textbf{Q2.} 
\textit{Does \alg{} invent mutually independent options?}
\textit{Can the options be composed and reused effectively?}

Options invented by \alg{} have a key advantage: their interpretable symbolic representation, where each option's initiation and termination conditions are defined in terms of specific value ranges of variables. 
Our analysis revealed that the invented options express distinct, complementary behaviors, with each option primarily affecting different state variables and value ranges. E.g., in the taxi domain, \alg{} invented four options that operate independently: two navigation options that affect different values of taxi location variable and specialize in moving to pickup/drop-off locations, and two passenger interaction options that affect different values of passenger variables and focus on picking up/dropping off the passenger. These options demonstrate mutual independence through minimal overlap in their core affected variables and value ranges in terminations of the options. Their clear symbolic endpoints enable direct chaining of options, making them both composable and reusable.

\section{Related Work}
 \label{sec:related}


Abstraction has been a topic of significant research interest \citep{karia2022learning,shah2024hierarchical,shah2024reals,karia2024epistemic}.
Early research in RL largely focused on hand-designed abstractions \citep{andre2002state, dietterich2000hierarchical}, with more recent frameworks also using high-level planning models or action hierarchies \citep{illanes2020symbolic, kokel2021reprel}. Typically, research on learning abstractions has focused on either state abstraction or action abstraction in isolation \citep{jonsson2000automated, wang2024building}.
A variety of methods have been developed for automatic discovery of subgoals or options, such as
identifying bottleneck states through graph-partitioning \citep{
menache2002q, 
csimcsek2007betweenness, 
machado2017laplacian}, clustering \citep{mannor2004dynamic}, and frequency-based \citep{mcgovern2001automatic, stolle2002learning} techniques. A large body of work learns hierarchies in which a high-level policy sets subgoals for a lower-level policy to achieve \citep{vezhnevets2017feudal, nachum2018data}. 

Most prior research in option discovery 
 focuses on control tasks with short horizons, often using dense rewards or distance metrics due to computational intractability
\citep{bacon2017option,  bagaria2021skill}. Much of this research is limited to single-task settings \citep{bagaria2020option,riemer2018learning}. 
Many recent methods learn a fixed, prespecified number of options and depend on learning a policy over options to use them \citep{bacon2017option,
machado2017eigenoption,
khetarpal2020options,
klissarov2021flexible}. 
In contrast, our work tackles a stream of long-horizon, sparse-reward tasks by continually learning generalizable options with abstract representations and planning over them.

\section{Conclusion and Future Work}

This paper presents a novel approach to continual RL based on autonomously learning and utilizing symbolic abstract options. This work assumes full observability and discrete actions. An interesting future research direction is to 
extend our approach to settings with continuous parameterized actions. 
Optimality is another good direction for future work.
\section{Acknowledgments}
We thank Shivanshu Verma for helping with an earlier version of this approach. This work was supported in part by NSF under grant IIS 2419809.

\bibliography{aaai25}

\end{document}